\documentclass[letterpaper, 10 pt, conference]{ieeeconf}  % Comment this line out if you need a4paper
\usepackage{subfig}

\IEEEoverridecommandlockouts % This command is only needed if you want to use the \thanks command

\usepackage{amsmath,amsfonts}
\usepackage{graphicx}
\usepackage{textcomp}
\usepackage{comment}

\usepackage{amsthm}
\usepackage[utf8]{inputenc}
\usepackage[T1]{fontenc}
\usepackage{enumerate}
\usepackage[font=scriptsize,labelformat=simple]{subcaption}

\usepackage{soul}

\usepackage{hyperref}
\hypersetup{
    colorlinks=true,
    citecolor=black,
    linkcolor=black,
    filecolor=black,      
    urlcolor=black,
}

\usepackage[style=ieee]{biblatex}
  
\AtBeginBibliography{\footnotesize}  %\small %\footnotesize %\scriptsize
\usepackage{siunitx}
\usepackage{multirow}
\usepackage{subfig}

\usepackage[ruled,vlined]{algorithm2e}
\DontPrintSemicolon

\usepackage{algpseudocode}

\usepackage{amssymb}    % For additional math symbols
\usepackage{mathtools}  % For more math tools
\usepackage{bm}         % For bold math symbols

\usepackage{accents}  % Add this in the preamble

\usepackage{colortbl}  % For table colors
\usepackage{array}     % For custom column width
\usepackage{adjustbox} % To fit table in text width
\usepackage{wasysym}  % For \lightning and \hourglass
\usepackage{xcolor}  % For colored circles
\usepackage{pifont}% http://ctan.org/pkg/pifont
\definecolor{gold}{HTML}{E5AE1C}
\definecolor{safe}{HTML}{4169E1}
\definecolor{unsafe}{HTML}{C41230}

\theoremstyle{definition}
\newtheorem{remark}{Remark}

\title{\LARGE \bf
LLA-MPC: Fast Adaptive Control for Autonomous Racing
}

\author{Maitham F. AL-Sunni \quad\quad Hassan Almubarak \quad\quad  Katherine Horng \quad\quad John M. Dolan% <-this % stops a space
\thanks{M. F. AL-Sunni is with the Electrical \& Computer Engineering Department, Carnegie Mellon University, Pittsburgh, PA,
USA. Email: \href{mailto:malsunni@andrew.cmu.edu}{\tt malsunni@andrew.cmu.edu}}
\thanks{H. Almubarak is with the School of Electrical and Computer Engineering, Georgia Institute of Technology, Atlanta, GA, USA. Email: \href{mailto:halmubarak@gatech.edu}{\tt halmubarak@gatech.edu}}%
\thanks{K. Horng is with the Mechanical Engineering Department, Carnegie Mellon University, Pittsburgh, PA,
USA. Email: \href{mailto:khorng@andrew.cmu.edu}{\tt khorng@andrew.cmu.edu}}
\thanks{J. M. Dolan is with the Robotics Institute, Carnegie Mellon University, Pittsburgh, PA, USA. Email: \href{mailto:jdolan@andrew.cmu.edu}{\tt jdolan@andrew.cmu.edu}}
\thanks{Our code, video presentation, and more details of our paper are available online at: \href{https://github.com/DRIVE-LAB-CMU/LLA-MPC}{https://github.com/DRIVE-LAB-CMU/LLA-MPC}}
}

\usepackage{siunitx}
\usepackage{multirow}

\usepackage[ruled,vlined]{algorithm2e}
\DontPrintSemicolon  % Uncomment this if you don't want semicolons at the end of each line

\usepackage{algpseudocode}

\usepackage{amsmath}    % For math symbols
\usepackage{amssymb}    % For additional math symbols
\usepackage{mathtools}  % For more math tools
\usepackage{bm}         % For bold math symbols

\usepackage{hyperref}   % For references

\usepackage{accents}  % Add this in the preamble

\newcommand{\ubar}[1]{\underaccent{\bar}{#1}}  % Define \ubar command

\DeclareMathOperator*{\argmin}{arg\,min}

\usepackage{tikz}
\usetikzlibrary{shapes, arrows, positioning}

\usepackage{graphicx}  % For including emojis
\usepackage{colortbl}  % For table colors
\usepackage{array}     % For custom column width
\usepackage{adjustbox} % To fit table in text width

\usepackage{pifont}  % For \ding{55} (X mark)
\usepackage{amssymb}  % For \checkmark
\usepackage{color}  % For colored circles
\usepackage{multirow}       % For multi-row cells
\usepackage{amssymb}% http://ctan.org/pkg/amssymb
\usepackage{pifont}% http://ctan.org/pkg/pifont

\definecolor{gold}{HTML}{E5AE1C}
\definecolor{safe}{HTML}{4169E1}
\definecolor{unsafe}{HTML}{C41230}
\usepackage{subfig}  % Use subfig instead of subcaption for IEEE format

\begin{document}

\maketitle
\thispagestyle{empty}
\pagestyle{empty}

\begin{abstract}
We present Look-Back and Look-Ahead Adaptive Model Predictive Control (LLA-MPC), a real-time adaptive control framework for autonomous racing that addresses the challenge of rapidly changing tire-surface interactions. Unlike existing approaches requiring substantial data collection or offline training, LLA-MPC employs a model bank for immediate adaptation without a learning period. It integrates two key mechanisms: a look-back window that evaluates recent vehicle behavior to select the most accurate model and a look-ahead horizon that optimizes trajectory planning based on the identified dynamics. The selected model and estimated friction coefficient are then incorporated into a trajectory planner to optimize reference paths in real-time. Experiments across diverse racing scenarios demonstrate that LLA-MPC outperforms state-of-the-art methods in adaptation speed and handling, even during sudden friction transitions. Its learning-free, computationally efficient design enables rapid adaptation, making it ideal for high-speed autonomous racing in multi-surface environments.
\end{abstract}

\section{Introduction}
\label{sec:intro}

High-performance autonomous racing requires vehicles to operate at the edge of handling limits while continuously adapting to dynamic and unpredictable conditions. A fundamental challenge in this domain is to sustain optimal performance despite variations in tire-road interactions, which significantly influence vehicle stability and maneuverability \cite{betz2022autonomous}. While Model Predictive Control (MPC) has proven effective for autonomous racing by explicitly considering system dynamics and constraints \cite{liniger2015optimization}, its performance heavily depends on the accuracy of the underlying vehicle model, particularly the tire dynamics.

Research in autonomous racing control has evolved along two primary approaches for addressing tire–surface interaction uncertainties. Off‐track identification methods, as exemplified by \cite{seong2023model,raji2022motion,voser2010analysis,becker2023model}, typically achieve high accuracy by relying on extensive offline data, specialized facilities, or large‐scale simulations; however, these methods are generally non-adaptive—they assume that once the vehicle model is identified, it remains fixed throughout the race. Even adaptive off‐track approaches, such as those based on meta-learning \cite{tsuchiya2024online,kalaria2025agile}, foundation models \cite{xiao2025anycar}, ensemble techniques \cite{nagy2023ensemblegaussianprocessesadaptive}, and GP‐based methods like BayesRace \cite{jain2021bayesrace}, require pre-race laps to collect substantial data and train their models, and often depend on large simulation datasets to cover a wide range of operating conditions. Moreover, the computational overhead associated with Gaussian Processes (even with sparse approximations) can hinder real-time prediction at high racing speeds.

In contrast, on-track identification methods update model parameters during the race. Traditional approaches rely on nonlinear least squares–based fitting \cite{rajamani2011vehicle,brunner2017repetitive}, which can be sensitive to noisy data, while optimization-based techniques \cite{bodmer2024optimization,kabzan2019learning} demand a reliable initial estimate of tire parameters. Among the adaptive on-track strategies, \cite{dikici2025learning} proposes an identification-only framework that addresses only mild model mismatches without an integrated control module, and \cite{kalaria2024adaptive} offers a full adaptive planning and control pipeline with friction estimation. However, the latter adapts slowly under abrupt condition shifts and has an average iteration time of approximately $0.06\mathrm{s}$, which poses challenges for high-speed racing. Additionally, many on-track methods depend on a preliminary data-collection interval $(10\text{–}60\mathrm{s})$, delaying full performance when conditions change suddenly mid-race.

\begin{figure}
\vspace{-5mm}
    \centering    \includegraphics[width=\linewidth]{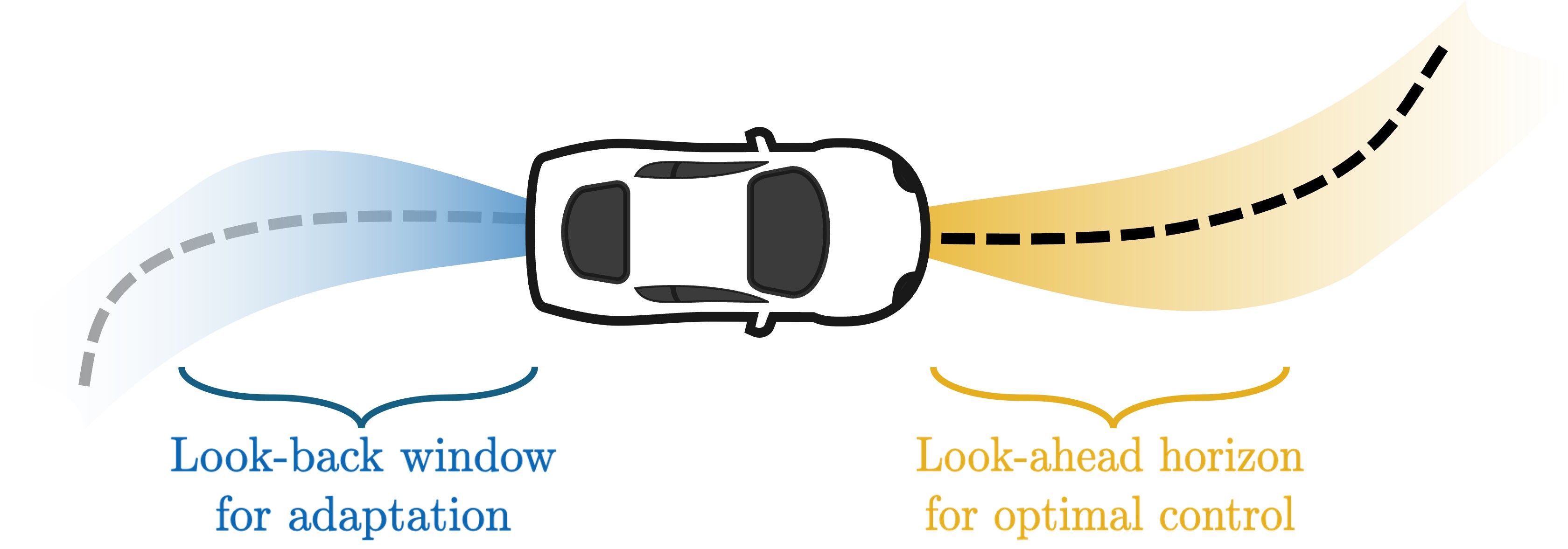}
    \caption{Illustration of the Look-Back and Look-Ahead Adaptive MPC (LLA-MPC) concept: The look-back window (blue) leverages past data for adaptation, while the look-ahead horizon (gold) optimizes future trajectory planning.}
    \label{fig:LLA-MPC-Cover}
    \vspace{-6mm}
\end{figure}

\begin{table}[h]
\vspace{-3mm}
\begin{center}
\caption{Comparison of adaptive on-track methods}
\label{tab:comparison}
\begin{tabular}{l|cccc}
\hline
Method & Training & Adaptation & Adaptive & Real-time \\
& Time & Speed & Planning & Capable \\
\hline
NNs-Based \cite{dikici2025learning} & {$\leq 60\mathrm{s}$} & {Slow} & {$\times$} & \textcolor{gold}{\checkmark} \\
ELM-based \cite{kalaria2024adaptive} & {$\leq 10\mathrm{s}$} & {Slow} & \textcolor{gold}{\checkmark} & \textcolor{gold}{\checkmark} \\
\textbf{LLA-MPC (Ours)} & \textcolor{gold}{\textbf{None}} & \textcolor{gold}{\textbf{Rapid}} & \textcolor{gold}{\checkmark} & \textcolor{gold}{\checkmark} \\
\hline
\end{tabular}
\end{center}
\vspace{-4mm}
\end{table}
Our work falls within the second approach—on-track identification—but overcomes the main limitations of existing methods. We propose {\it {\underline L}ook-{B}ack and {\underline L}ook-{A}head {\underline A}daptive {\underline M}odel {\underline P}redictive {\underline C}ontrol (LLA-MPC)}, a real-time framework requiring no initial learning laps or dedicated data-collection phase. Inspired by the general concept of multiple-model adaptive control \cite{narendra1997adaptive}, LLA-MPC assembles a large bank of physics-based models covering varied tire-surface interactions, continuously selects the best-fit model from a short {\it look-back} window, then uses that model for an adaptive, {\it look-ahead} MPC. Crucially, we estimate the friction coefficient of the road and pass it to the planner to determine appropriate speeds for the current road conditions, similar in principle to \cite{kalaria2024adaptive} but realized via a learning-free multi-model bank rather than a neural network. As a result, LLA-MPC enables immediate adaptation to both harsh gradual and sudden friction changes from the start of a race, negating the latency or extensive pre-race data needed by prior methods. \autoref{tab:comparison} summarizes the comparisons between our approach and the closest adaptive on-track methods in the literature.

The contributions of our work are threefold:  
\begin{enumerate}
    \item A learning-free adaptive MPC framework that achieves rapid adaptation through parallel model banking that operates in real time without the need of data collection.
    \item We use our learning-free approach to integrate friction estimation for adaptive trajectory planning by accurately estimating the friction coefficient of the road and passing it to the path planner.
    \item We test our proposed method on a variety of challenging scenarios. The results suggest that our method outperforms state-of-the-art adaptive on-track approaches in lap times, safety, tracking performance, and computational cost.
\end{enumerate}

We demonstrate through extensive experiments that our approach maintains consistent tracking performance during both gradual and sudden friction changes while existing methods struggle with rapid transitions. Results show successful adaptation across various racing scenarios without compromising vehicle stability or performance.

The remainder of this paper is organized as follows: Section \ref{sec:lla-mpc} details the LLA-MPC framework and its theoretical properties. Section \ref{sec:auto_race} describes the application to autonomous racing. Section \ref{sec:results} presents experimental results and comparisons. Finally, Section \ref{sec:conclusion} concludes with a discussion of future work.

\section{Look-Back and Look-Ahead Adaptive MPC}
\label{sec:lla-mpc}
\subsection{Problem Setting}
Consider the discrete-time dynamical system
\begin{align}
    \label{eq:unknown_dynamics}
    \mathbf{x}_{k+1} = f^\star(\mathbf{x}_k,\mathbf{u}_k),
\end{align}
where $\mathbf{x}_k \in \mathbb{R}^n$ and $\mathbf{u}_k \in \mathbb{R}^m$ are the state and the control input at the time step $k$, and $f^\star : \mathbb{R}^n \times \mathbb{R}^m \to \mathbb{R}^n$ is the dynamics function. 

For many applications, researchers and engineers have successfully created parametric models that can accurately capture the dynamics of the unknown system \eqref{eq:unknown_dynamics}. Classical system identification approaches are usually used to identify the parameters of the parametric models that give the best identification accuracy. However, such methods require a lot of data to converge, often require offline computations/learning, and lack the feature of online adaptation. Putting the last part in other words, the identification process assumes that the properties of the real system \eqref{eq:unknown_dynamics} do not change while it is running. This leads to having fixed learned parameters that cannot adapt to changes in the real system when it is used for control purposes. 

Our aim is to design a control framework for the unknown system \eqref{eq:unknown_dynamics} based on an identification method that possesses the following features: 1) learning-free, i.e., does not require data collection, 2) onboard-deployable, i.e., no offline efforts and operates in real-time, 3) adapts rapidly to changes in the system and the environment. 

\subsection{Methodology}
Inspired by the model path integral control framework that turned the classical MPC problem into a much computationally cheaper process using just sampling, we take sampling as the spirit of our approach to develop the method of identification we aim for.

Let
\begin{align}
    \label{eq:model_structure}
    \mathbf{x}_{k+1} = f_s(\mathbf{x}_k,\mathbf{u}_k;\pmb{\theta}_s)
\end{align}
be a parametric model structure (i.e., functional form)\footnote{As an example, in autonomous racing, one could have a kinematic model, or a dynamic model as a model structure.} $s$ that can potentially model the unknown dynamics \eqref{eq:unknown_dynamics} with $\pmb{\theta}_s \in \mathbb{R}^{p_s}$ being the set of parameters specific to that structure. Here, $p_s$ represents the number of parameters for model structure $s$. % \textcolor{red}{Note that changes in the the system or its environment may lead to a need of adapting $\pmb{\theta}_s$.}

For many applications, the parameters in $\pmb{\theta}_s$ are not completely unknown, i.e., their values lie in some known range. Specifically, for each parameter $\theta_{s,i}$ (the $i$-th parameter of model structure $s$), we know that $\theta_{s,i} \in [\ubar{\theta}_{s,i}, \bar{\theta}_{s,i}]$ for all $i \in \{1, 2, ..., p_s\}$. In vector notation, we have $\ubar{\pmb{\theta}}_s \leq \pmb{\theta}_s \leq \bar{\pmb{\theta}}_s$ where $\ubar{\pmb{\theta}}_s$ and $\bar{\pmb{\theta}}_s$ represent the vectors for the lower and upper bounds, respectively.

With this, we can form a bank of models that incorporates multiple structures:
\begin{align}
\mathcal{F} = \mathcal{F}_1 \cup \mathcal{F}_2 \cup \cdots \cup \mathcal{F}_S,
\end{align}
where $S$ is the total number of different model structures, and each $\mathcal{F}_s = \{f_s(\cdot; \pmb{\theta}_s^1), f_s(\cdot; \pmb{\theta}_s^2), \ldots, f_s(\cdot; \pmb{\theta}_s^{N_s})\}$ contains $N_s$ models with the same structure $s$ but different parameter values sampled uniformly\footnote{If nominal or usual values of the parameters are known, one could use a normal distribution centered at the nominal values. However, in the case where we have a changing system, we need the bank to be diverse enough to capture different conditions.} (entry-wise) from the known ranges, i.e.,
\begin{align}    
\label{eq:sampling}
\pmb{\theta}_s^j \sim \text{Uniform}(\ubar{\pmb{\theta}}_s,\bar{\pmb{\theta}}_s), \ \forall j \in \{1, 2, 3, \hdots N_s\}.
\end{align}
\begin{remark}
    \label{remark:as N grows}
    When the total number of models in the bank $N = \sum_{s=1}^S N_s$  grows, the likelihood of having a model that is close to the actual unknown system increases. The exact likelihood would depend on the number of parameters each model structure has and how wide the ranges for these parameters are. 
\end{remark}

\begin{figure}
    \centering
    \includegraphics[width=\linewidth]{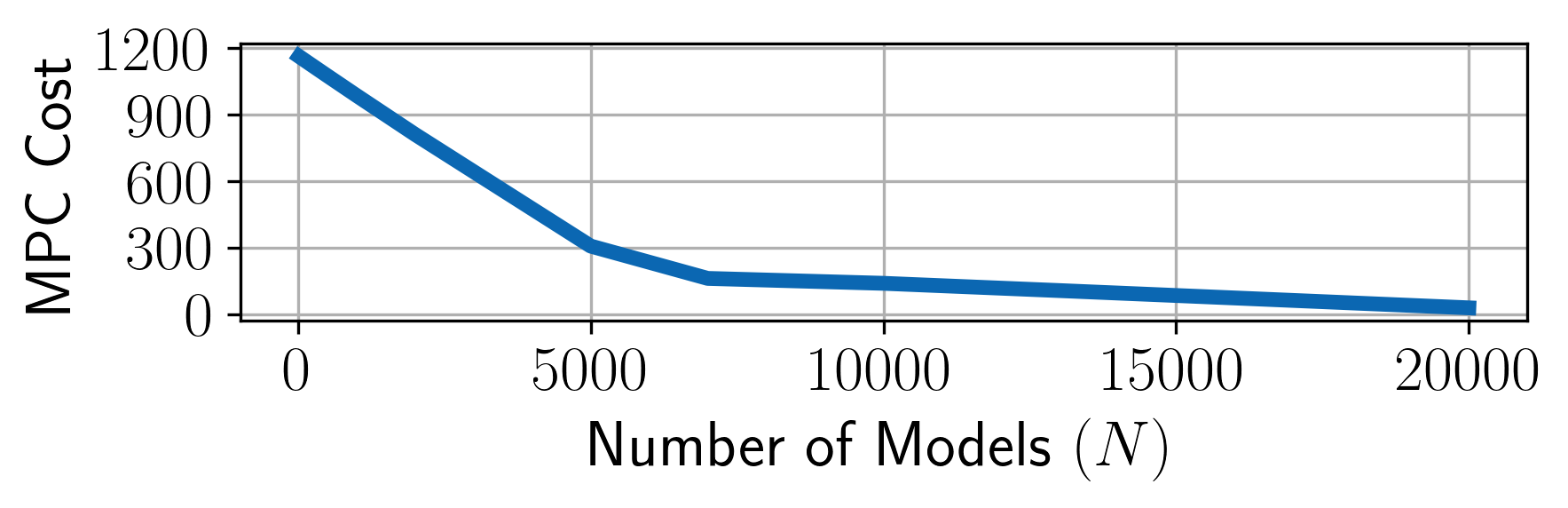}
    \caption{Impact of model bank size ($N$) on LLA-MPC performance. A larger $N$ reduces the MPC cost, improving adaptation and control.}
    \label{fig:performance as N increases}
    \vspace{-7mm}
\end{figure}

\autoref{fig:performance as N increases} illustrates the effect of the size of the model bank ($N$) on LLA-MPC performance for an autonomous racing car trying to complete $3$ laps with progressive tire degradation, an experiment we perform in \autoref{sec:results}. As $N$ increases, the MPC cost decreases, indicating improved adaptation and control. This highlights the advantage of having a diverse set of models for rapid and accurate response to changing conditions.

Given a sufficiently rich bank of models, we aim to design a control methodology that can dynamically adapt to changes in the environment by choosing the most accurate model for planning and control. For this, we propose LLA-MPC.

\subsubsection{Look-Back}
Our idea leverages parallel computing to simultaneously predict system evolution and evaluate the one-step-ahead predictive accuracy of candidate models at each time step. At every step, a \textit{look-back} over the past $W$ time steps is performed, selecting the model that minimizes the accumulated squared error as the new model for MPC. More specifically, at every time step $k$, we compute the squared error between the state $\mathbf{x}_{k}$ and the single-step predicted state for all candidate models in the bank as
\begin{align}
    \label{eq:error_window_small}
    e_k^j = \left|\left|\mathbf{x}_{k} - f^j(\mathbf{x}_{k-1},\mathbf{u}_{k-1})\right|\right|^2, \ \forall j \in \{1,\hdots,N\},
\end{align}
where each $f^j\sim\mathcal{F}$ is a sample in the bank. For notational convenience, we drop the parameter $\theta$ from the different $f^j$ models in the bank. It is worth noting that propagating all models to collect their predictions as well as computing the MSE are fully computationally parallelizable. For every time step $k \geq W$, we \textit{look-back} and compute the accumulated single-step prediction error
\begin{align}
    \label{eq:error_window}
    \mathcal{E}_k^j = \sum_{i=0}^{W-1} e^j_{k-i}, \ \forall j \in \{1,\hdots,N\}.
\end{align}

Finally, we select the model that minimizes the accumulated MSE by solving
\begin{align} \label{eq:choose_model}
    j_k^{\star} = \argmin_{j} \mathcal{E}_k^j,
\end{align}
which provides us with the model that best predicts the actual system at time step $k$.

\begin{remark} \label{remark: multi-structure}
The error computation in \eqref{eq:error_window} only requires that each model structure can predict the same observable state vector, but places no constraints on the internal structure or complexity of the model structures themselves. This flexibility allows our framework to incorporate models ranging from simple kinematic approximations to complex dynamic formulations, or even data-driven models. The only requirement is that each model must map the same input space to the same output space to enable fair comparison through the error metric.
\end{remark}

\textbf{Sliding Window Selection for Error Evaluation:}
An important question here is how to select the look-back window $W$, which plays a crucial role in balancing adaptation speed and accuracy in model selection. Memory-less techniques often lead to abrupt model switching, which can increase estimation error and induce chattering in control systems, ultimately degrading performance \cite{maybeck1982stochastic}. On the other hand, larger memory models provide smoother transitions but incur computational complexity and slower adaptation, especially in the presence of sudden system changes. 

\begin{figure}
    \centering
    \includegraphics[width=\linewidth]{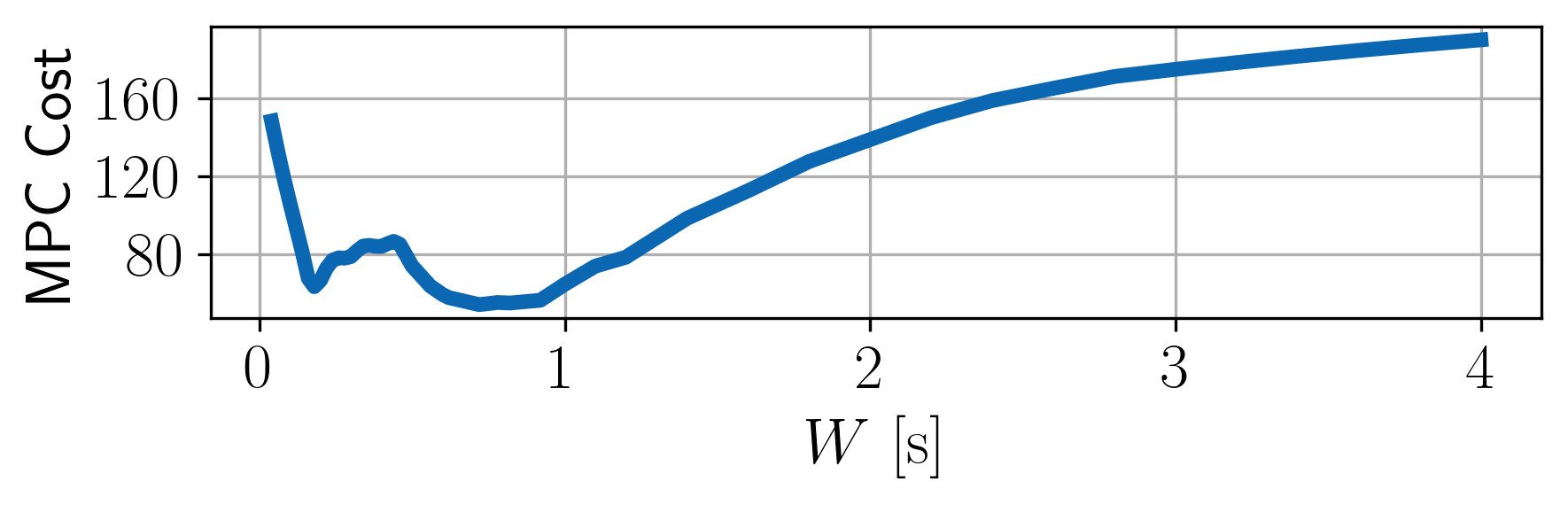}
    \caption{Impact of size of the look-back window ($W$) when $N=20000$. A balanced $W$ reduces the MPC cost.}
    \label{fig:w_effect}
        \vspace{-7mm}
\end{figure}

In our approach, a smaller $W$ enables faster adaptation but increases sensitivity to noise and outliers, while a larger $W$ results in slower adaptation, albeit smoother transitions. However, we still choose a sufficiently small window $W$ because a diverse and large model bank mitigates the impact of noise and outliers. While a large model bank introduces higher computational complexity, parallel computing tools significantly reduce this cost. This is particularly important for real-time applications like racing, where fast adaptation is essential. Using a limited look-back window $W$ rather than considering the entire history allows the model selection process to remain computationally efficient while staying responsive to changes in the system. \autoref{fig:w_effect} shows the effect of the choice of $W$ for an autonomous racing car trying to complete $3$ laps with progressive tire degradation, an experiment we perform in \autoref{sec:results}. A small yet large enough $W$ gives the best performance.

\subsubsection{Look-Ahead}
The next step in our proposed scheme involves implementing a nonlinear model predictive controller using the selected $j_k^\star$th model\footnote{For the very short period of time when $k<W$, we run the system using a controller that uses a nominal model $f_{\text{nom}}$.}, $f^{j_k^\star}$, at each time step $k$ to compute the sequence of control inputs over the prediction horizon. At each subsequent time step $k+1$, we re-evaluate the error window as defined in \eqref{eq:error_window} and solve \eqref{eq:choose_model} to determine the best-fitting model to be adopted to solve the MPC problem. Given the state estimate $\mathbf{x}_t$ at time $t$, the MPC formulation is:
\begin{align}
    \label{eq:gen_mpc}
    \min_{{\mathbf{u}}_0, \dots, {\mathbf{u}}_{H-1}} & \sum_{k=0}^{H-1} \ell(\mathbf{x}_k, \mathbf{u}_k) + {\ell_T}(\mathbf{x}_H), \\
    \text{s.t.} \quad & \mathbf{x}_{k+1} = f^{j_k^\star}(\mathbf{x}_k, \mathbf{u}_k), \quad \mathbf{x}_0 = \mathbf{x}_t, \nonumber \\
    & \mathbf{x}_k \in \mathcal{X}, \quad \mathbf{u}_k \in \mathcal{U},   \nonumber
\end{align}
where $H$ is the \textit{look-ahead} horizon, a.k.a. the prediction horizon, $\ell$ and $\ell_T$ are the nominal running and terminal cost functions respectively, $\mathcal{X}$ is some desired subset in the state space, and $\mathcal{U}$ is the set of admissible controls. Algorithm \ref{Alg:LLA-MPC} illustrates and provides the details of LLA-MPC. It is important to note that the rolling sum of the MSE in our algorithm requires only $\mathcal{O}(N)$ memory when implemented efficiently in a rolling fashion.

\begin{algorithm}[t]
\caption{LLA-MPC}
\label{Alg:LLA-MPC}
% \DontPrintSemicolon

\KwIn{
    Initial state $\mathbf{x}_0$, Nominal model $f_{\text{nom}}$, Model bank $\mathcal{F}$ with $N$ models, look-back window $W$, look-ahead horizon $H$, cost functions $\ell$ and $\ell_T$, final time $T$, sampling time $\Delta T$
}
\KwOut{
    Adaptively Optimized trajectory
}

\For{$k=0$ \textbf{to} $\frac{T}{\Delta T}$}{
    
    \If{k < W}{
    $f \gets f_{\text{nom}}$ \tcp{Use nominal model}
    }
    \Else{
    $\mathcal{E}_k^j = \sum_{i=0}^{W-1} e^j_{k-i}, \ \forall j \in \{1,\hdots N\}$\;
    $j_k^{\star} = \argmin_{j} \mathcal{E}_k^j$ \tcp{\autoref{eq:choose_model}}
    $f \gets f^{j_k^\star}$ \tcp{Use best model}
    
    }
    $\mathbf{x}_{\text{ref}} = {\tt planner}(\mathbf{x}_k)$\;
    $\mathbf{u}_{k} \gets {\tt MPC}(f, \mathbf{x}_k, \mathbf{x}_{\text{ref}})$ \tcp{\autoref{eq:gen_mpc}}
    $\mathbf{x}_{k+1} = {\tt dynamics}(\mathbf{x}_k, \mathbf{u}_k)$ \\
    $e_{k+1}^j= \left|\left|\mathbf{x}_{k} - f^j(\mathbf{x}_{k-1},\mathbf{u}_{k-1})\right|\right|^2 \ \forall j \in \{1,\hdots N\}$ \tcp{\autoref{eq:error_window_small}}
    }
\end{algorithm}

%\begin{remark}
%It should be noted that our algorithm requires no data storage as we can implement the lock-back window error in a rolling fashion. Even when this is not done, the worst implementation for Algorithm \ref{Alg:LLA-MPC} requires saving the errors for the last $W$ time steps only.
%\end{remark}

% \textcolor{red}{add the block diagram... }
% \begin{figure}[h]
%     \centering
%     \includegraphics[width=\linewidth]{figures/LLA_mpc_black_diag.png}
%     \caption{Overview of the Look-Back and Look-Ahead Adaptive MPC (LLA-MPC) framework. The system integrates a model-predictive controller, a path planner, and an adaptive model selection mechanism to enhance performance in autonomous racing.}
%     \label{fig:LLA-MPC}
% \end{figure}

\section{Application: Autonomous Racing}
\label{sec:auto_race}

We propose that LLA-MPC is an excellent approach for autonomous racing, where fast adaptation and precise control are essential for optimal performance. By utilizing LLA-MPC, we aim to effectively handle the dynamic and high-speed environment of racing, ensuring that model selection and control inputs are rapidly adjusted to meet the continuously changing conditions on the track.

\subsection{Vehicle Model and Parameters}

We adopt the dynamic bicycle model (DBM), a widely known descriptive model of autonomous vehicles. The DBM captures the essential vehicle dynamics while remaining computationally tractable for real-time control. It must be noted, however, that this model includes vehicle and tire parameters that can rapidly change over time in autonomous racing \cite{liniger2015optimization,rosolia2019learning,hewing2018cautious}. 

The model can be written as
\begin{align}
\underbrace{\begin{bmatrix}
    \dot{x} \\ \dot{y} \\ \dot{\phi} \\ \dot{v}_x \\ \dot{v}_y \\ \dot{\omega} \\ \dot{\delta}
\end{bmatrix}}_{\dot{\mathbf{x}}}
=
\underbrace{\begin{bmatrix}
    v_x \cos(\phi) - v_y \sin(\phi) \\
    v_x \sin(\phi) + v_y \cos(\phi) \\
    \omega \\
    \frac{1}{m} \left( F_{r,x} - F_{\mathrm{f},y} \sin(\delta) + m v_y \omega - mg \sin(p) \right) \\
    \frac{1}{m} \left( F_{r,y} + F_{\mathrm{f},y} \cos(\delta) - m v_x \omega + mg \sin(r) \right) \\
    \frac{1}{I_z} \left( F_{\mathrm{f},y} l_\mathrm{f} \cos(\delta) - F_{r,y} l_r \right) \\
    \frac{\Delta \delta}{\Delta t}
\end{bmatrix}}_{f_{\text{DBM}}(\mathbf{x},\mathbf{u})}, \nonumber
\end{align}
where $(x,y)$ is the location in an inertial frame, $\phi$ is the inertial heading, $v_x$ and $v_y$ are the speeds in the body frame, $\omega$ is the angular velocity, $\delta$ is the steering angle, $F_{r,x}$ is the longitudinal force in the body frame, $F_{\mathrm{f},y}$ and $F_{r,y}$ are the lateral forces in the body frame\footnote{The subscripts $\mathrm{f}$ and $r$ denotes front and rear wheels.}, $g$ is the
gravitational constant, $m$ is the mass, $p$ and $r$ are the roll and pitch angles of the vehicle, $I_z$ is the moment of inertia in the vertical direction about the center of mass (CoM), $l_\mathrm{f}$ and $l_r$ are the distances from the CoM to the front and rear wheels respectively, $\Delta\delta$ is the change in the steering angle, $\Delta t$ is the sampling time. In the DBM model, the longitudinal force is given by $F_{r,x} = \left( C_{m_1} - C_{m_2} v_x \right) d - C_{ro} - C_d v_x^2$, where $C_{m_1}$ and $C_{m_2}$ are known parameters for the DC motor of the vehicle, $C_{ro}$ is the rolling resistance, and $C_d$ is the aerodynamic force constant. The lateral forces are usually modeled using a Pacejka tire model \cite{bakker1987tyre} as follows
\begin{align}
    F_{\mathrm{f},y} &= D_\mathrm{f} \sin \left( C_\mathrm{f} \arctan \left( B_\mathrm{f} \alpha_\mathrm{f} \right) \right),  \\
    F_{r,y} &= D_r \sin \left( C_r \arctan \left( B_r \alpha_r \right) \right),
\end{align}
where $\alpha_\mathrm{f} = \delta - \arctan \left( \frac{\omega l_\mathrm{f} + v_y}{v_x} \right)$ and $ 
    \alpha_r = \arctan \left( \frac{\omega l_r - v_y}{v_x} \right)$ are the slip angles and $B_\mathrm{f}$, $B_r$, $C_\mathrm{f}$, $C_r$, $D_\mathrm{f}$, $D_r$ are the tire parameters which are tire and track specific. 

Such vehicle and tire parameters need to be updated dynamically as tire wear, temperature, track conditions, load shifts, and driving maneuvers change rapidly, ensuring accurate modeling for optimal performance, stability, and control in autonomous racing. Consequently, \textit{on-track} identification of those parameters is a challenging problem that has been a focus in the recent literature \cite{dikici2025learning, kalaria2024adaptive,rosolia2019learning}. To address this challenge, we focus on the dynamic identification of key parameters essential for accurate vehicle modeling and control. Specifically, we consider the real-time adaptation of the following parameters  $\pmb{\theta} = [B_\mathrm{f}, B_r, C_\mathrm{f}, C_r, D_\mathrm{f}, D_r, C_{ro}, C_d]$. Other parameters, such as the mass $m$, moment of inertia $I_z$, and geometric properties $l_\mathrm{f}, l_r$, are excluded from  $\pmb{\theta}$ as they are usually directly measured or determined with high confidence, making their online adaptation unnecessary.  

As noted in \autoref{remark: multi-structure}, our LLA-MPC framework supports diverse model structures, but for autonomous racing, it suffices that our model bank $\mathcal{F}$ consists of a dynamic bicycle model structure only due to its effectiveness in capturing key dynamics. Given the vehicle's general characteristics, reasonable bounds for the parameters $\pmb{\theta}$ can be established, denoted as $\ubar{\pmb{\theta}} = [\ubar{B}_{\mathrm{f}}, \ubar{B}_r, \ubar{C}_{\mathrm{f}}, \ubar{C}_r, \ubar{D}_{\mathrm{f}}, \ubar{D}_r, \ubar{C}_{\mathrm{ro}}, \ubar{C}_d]$ and $\bar{\pmb{\theta}} = [\bar{B}_{\mathrm{f}}, \bar{B}_r, \bar{C}_{\mathrm{f}}, \bar{C}_r, \bar{D}_{\mathrm{f}}, \bar{D}_r, \bar{C}_{\mathrm{ro}}, \bar{C}_d]$, within which the true parameter values are expected to lie. These bounds facilitate the construction of a model bank containing $N$ dynamic bicycle models, $\mathcal{F} = \{f_1(\cdot; \pmb{\theta}^1), f_2(\cdot; \pmb{\theta}^2), \ldots, f_N(\cdot; \pmb{\theta}^N)\}$, by uniformly sampling the parameter space, as detailed in equation \eqref{eq:sampling}. Our approach allows for comprehensive coverage of a wide range of bounds, enabling the system to adapt effectively to varying operational conditions. When the best-fitting model $j_k^{\star}$ is selected at time step $k$ using \eqref{eq:choose_model}, we are effectively selecting the parameter set $\pmb{\theta}^{j_k^{\star}}$ that best describes the current vehicle-surface interaction. These selected parameters are then used for state prediction in the MPC formulation and for adapting the friction coefficient that is essential for dynamic and agile path planning, as detailed in the next subsection.

\subsection{Racing Line Generation and Adaptive Planning}
\label{sec:raceline}

In high-performance driving, such as autonomous racing, the road’s friction coefficient significantly influences the vehicle's handling capabilities. A reference speed that fails to account for this varying friction can lead to suboptimal performance, either underutilizing the vehicle's potential or risking loss of traction. By estimating the friction coefficient using peak lateral force parameters, derived from real-time tire parameter estimates, we can dynamically adapt the reference speed to the current road conditions. This adaptive strategy ensures that the vehicle operates within safe limits while maximizing performance, optimizing both safety and handling. Real-time estimation of road friction, combined with dynamic speed adjustment, enables the vehicle to intelligently respond to changing track conditions, providing both stability and efficiency in its path planning.

To begin, we generate an optimal racing line using the minimum-curvature approach \cite{heilmeier2020minimum}, where the track is represented as a sequence of points with associated width. This racing line serves as the spatial foundation for trajectory planning. Next, we adopt the method from \cite{kalaria2024adaptive} to create a library of velocity profiles optimized for different friction coefficients $\mu$ within the range $[\mu_{\min}, \mu_{\max}]$. Each velocity profile in this library corresponds to a specific friction coefficient, providing appropriate speed references along the racing line. This step allows for efficient online adaptation without the need for real-time optimization of speed profiles.

Our approach integrates on-track adaptive identification, model predictive control, and adaptive trajectory planning to optimize vehicle performance under rapidly changing race conditions. Due to possible fast switching of friction values, rapid changes in road conditions can introduce significant short-term fluctuations in the friction estimate. To mitigate these fluctuations and ensure the stability of the estimate, a smoothing technique is applied. First, the peak lateral forces of the selected model $f^{j^*_k}$ are used to compute the friction coefficient normalized by the vehicle's weight as
\begin{align}
    \bar{\mu}_k = \frac{D_{\mathrm{f}}^{j_k^\star} + D_{r}^{j_k^\star} }{2 m g}.
\end{align}
To improve the stability and consistency of the estimate, exponential smoothing is applied with $\gamma \in (0,1)$, acting as a low-pass filter that allows for rapid adaptation while reducing the impact of outliers:
\begin{equation}
    \mu_k = \gamma \bar{\mu}_k + (1 - \gamma) \mu_{k-1}.
\end{equation}
This procedure ensures that the friction estimate remains robust and responsive, forming a reliable foundation for dynamically adjusting the reference speed in real-time to accommodate varying road conditions.

Finally, we use the estimated $\mu$ to select, or interpolate, between velocity profiles from the library. Unlike \cite{kalaria2024adaptive}, which relies on neural networks requiring extensive training, our model bank provides immediate friction estimates, enabling rapid adaptation. This integration allows the system to respond to sudden friction changes without the need of data collection that might compromise safety or stability, maintaining performance even during rapid surface transitions.

\section{Results}
\label{sec:results}
We perform various experiments on two different simulation platforms, 1) the 1:43 numeric simulator \cite{liniger2015optimization} and 2) the high-fidelity simulator CARLA \cite{dosovitskiy2017carla}.

\begin{figure}
    \centering    \includegraphics[width=\linewidth]{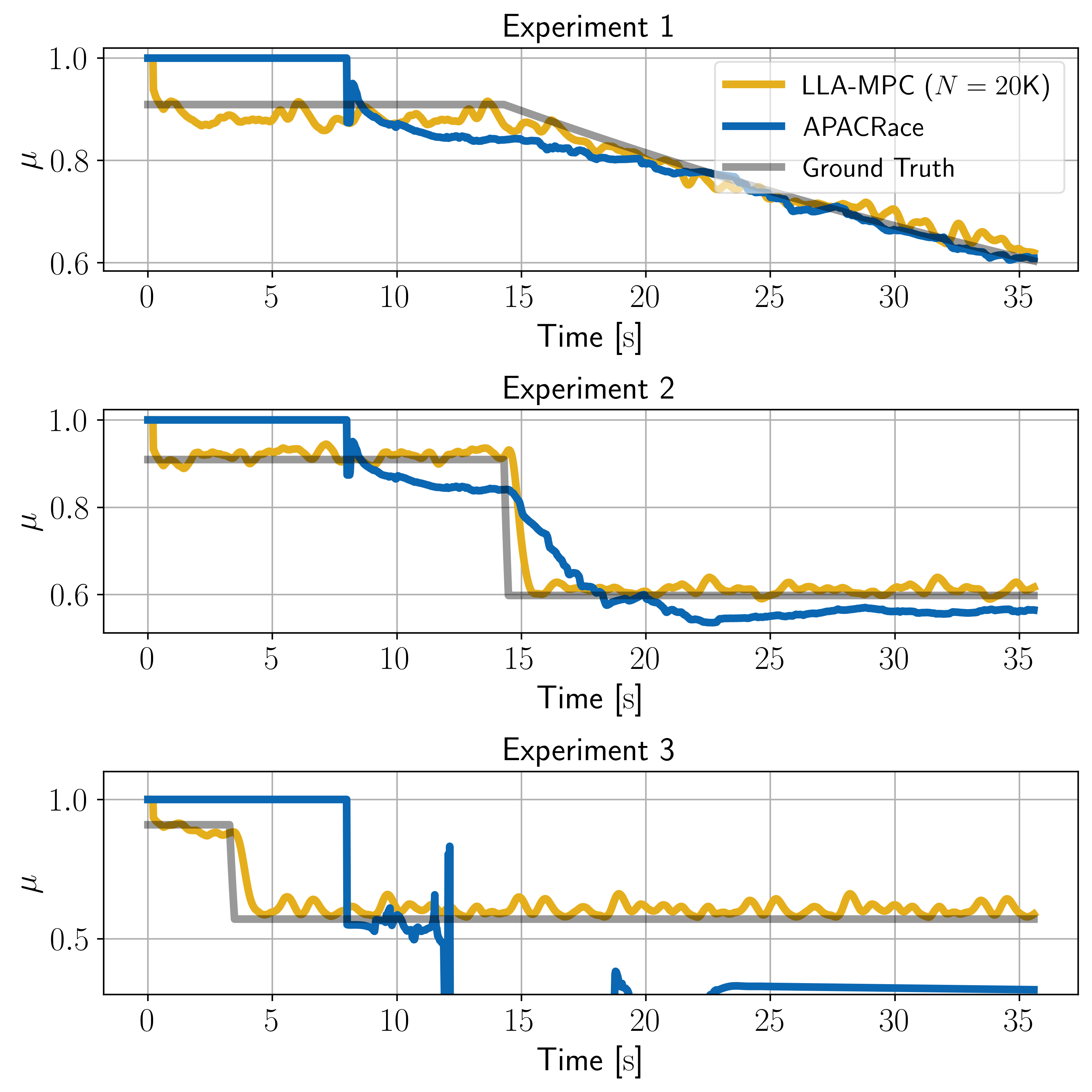}
    \caption{Comparison of friction coefficient ($\mu$) estimations over time for the ETHZ track.}
    \label{fig:all mus}
    \vspace{-5mm}
\end{figure}

\subsection{Numerical Simulations}
In the 1:43 numeric simulator \cite{liniger2015optimization}, we validate our approach on the ETHZ \cite{liniger2015optimization} and ETHZMobil \cite{jain2021bayesrace} tracks. In the simulator, typical values are given for the regular vehicle's parameters. Since our framework assumes no knowledge of these parameters' values and only assumes some knowledge of their ranges, we choose a range that starts from a lower bound $\ubar{\pmb{\theta}}$ that is $-\%150$ of the actual parameters, and ends with an upper bound $\bar{\pmb{\theta}}$ that is $+\%150$ of the actual parameters. For the number of models in the bank, we consider both $N = 20000$ and $N=10000$ where every model $j$ from the $N$ models is sampled from $\pmb{\theta}^j \sim \text{Uniform}(\ubar{\pmb{\theta}},\bar{\pmb{\theta}})$ as described in \eqref{eq:sampling}. Thanks to our parallelized implementation, we can evaluate this large model bank in real-time, enabling fine-grained coverage of the parameter space. Also, we choose the look-back window as $W=0.2 \mathrm{s}$ and the look-ahead horizon as $H=0.4 \mathrm{s}$ with sampling time $\Delta T = 0.02 \mathrm{s}$. To make diverse and fair comparisons between the different approaches, we cover three different cases of changing environments, as detailed below. For the three experiments, we will limit the comparisons to be against APACRace \cite{kalaria2024adaptive} as it is, to the best of our knowledge, the fastest (in both training and adaptation) on-track identification method in the literature. Additionally, we compare both methods against an Oracle, which is an MPC controller that instantaneously knows all changes in the environment. This Oracle serves as the ultimate best performance. For all tracks and methods, we let the racing vehicle complete 3 laps. The metrics that we consider for comparisons are 1) lap time, 2) violation time, which is the time that the vehicle spent outside the bounds of the racing track, and 3) mean deviation from the race line,  which is the minimum-curvature path that was discussed in Section \ref{sec:raceline}. For all experiments, both algorithms start with an initial estimated friction coefficient $\mu=1$. \autoref{table: ETHZ Track} and \autoref{table: ETHZMobil Track} summarize the comparisons between the different approaches on both tracks in different racing conditions (numbers are averaged over $10$ runs). In the table, we can see that with $N=10000$ LLA-MPC performs generally worse than when $N=20000$. This realizes the insight presented in \autoref{remark:as N grows} and \autoref{fig:performance as N increases}. It is worth noting that per-iteration computation time of LLA-MPC takes only $0.01\mathrm{s}$ extra than the Oracle, whereas APACRace requires double LLA-MPC's computation times under the same conditions.

\begin{table}
%\scriptsize
% \footnotesize
\caption{ETHZ Track}
\centering
\scriptsize % This sets the font size for the table
 \begin{tabular}{l | c | c | c | c } 
   & \multirow{2}{*}{\textcolor{gold}{\centering Oracle}}&  \multirow{2}{1.1cm}{\centering LLA-MPC $N=20$K} & \multirow{2}{1.1cm}{\centering LLA-MPC $N=10$K} & \multirow{2}{*}{\centering APACRace}\\ 
    & &  & & \\ 
 \hline
   \multicolumn{5}{c}{\bf Experiment 1} \\ 
 \hline
 Lap 1 Time ($\mathrm{s}$)       & ${\color{gold}{7.68}}$  & $ \mathbf{7.76} $  & $8.01$   & $8.54$ \\ 
 Lap 2 Time ($\mathrm{s}$)       & ${\color{gold}{7.50}}$  & $\mathbf{7.58} $   & $7.85 $   & $7.65$  \\ 
 Lap 3 Time ($\mathrm{s}$)       & ${\color{gold}{7.76}}$  & $\mathbf{7.92} $   & $ 8.14$   & $7.96$ \\ 
 Violation Time ($\mathrm{s}$)   & ${\color{gold}{0.84}}$  & $ \mathbf{0.78} $   & $1.07$  & $1.24$  \\
 Mean Deviation ($\mathrm{m}$)   & ${\color{gold}{0.04}}$  & $ \mathbf{0.04} $  & $0.05$   & $0.05$  \\
 \hline
 \multicolumn{5}{c}{\bf Experiment 2} \\ 
 \hline
 Lap 1 Time ($\mathrm{s}$)   & ${\color{gold}{7.68}}$ & $\mathbf{7.74}$   & $7.96$     & $8.54$ \\ 
 Lap 2 Time ($\mathrm{s}$)   & ${\color{gold}{7.62}}$  & $\mathbf{7.76}$    & $7.96$    & $7.93$ \\ 
 Lap 3 Time ($\mathrm{s}$)   & ${\color{gold}{9.00}}$  & $\mathbf{9.16}$     & $ 9.22$   & $9.36$ \\ 
 Violation Time ($\mathrm{s}$)&${\color{gold}{0.72}}$  & $\mathbf{0.65}$    & $1.04$    & $2.12$ \\
 Mean Deviation ($\mathrm{m}$)&${\color{gold}{0.04}}$  &  $\mathbf{0.04}$   & $\mathbf{0.04}$    &  $0.05$  \\
 \hline
 \multicolumn{5}{c}{\bf Experiment 3} \\  
 \hline
 Lap 1 Time ($\mathrm{s}$)  & ${\color{gold}{8.60}}$  & $\mathbf{8.74}$   & $9.00$    & $12.98$ \\ 
 Lap 2 Time ($\mathrm{s}$)  & ${\color{gold}{9.22}}$   &  $\mathbf{9.26}$   & $9.40$    & $\times$ \\ 
 Lap 3 Time ($\mathrm{s}$)  & ${\color{gold}{9.21}}$   & $\mathbf{9.28}$   & $9.40$    & $\times$ \\ 
 Violation Time ($\mathrm{s}$)&${\color{gold}{0.48}}$  & $\mathbf{0.48}$ & $1.04$       & $\times$\\
 Mean Deviation ($\mathrm{m}$)&${\color{gold}{0.04}}$   &  $\mathbf{0.04}$  & $0.05$      & $\times$ \\
 \hline
  {\bf Avg. Comput. Time} ($\mathrm{s}$)&${\textcolor{gold}{0.02}}$   &  $\mathbf{0.03}$        & $\mathbf{0.03}$ & $0.06$ \\
  \hline
 \end{tabular}
 \label{table: ETHZ Track}
 % \vspace{-4mm}
\end{table}

\begin{table}
% \scriptsiz
% \footnotesize
\caption{ETHZMobil Track}
\centering
\scriptsize % This sets the font size for the table
 \begin{tabular}{l | c |c |c | c} 
   & \multirow{2}{*}{\textcolor{gold}{\centering Oracle}}&  \multirow{2}{1.1cm}{\centering LLA-MPC $N=20$K} & \multirow{2}{1.1cm}{\centering LLA-MPC $N=10$K} & \multirow{2}{*}{\centering APACRace}\\ 
    & &  & & \\ 
 \hline
   \multicolumn{5}{c}{\bf Experiment 1} \\ 
 \hline
 Lap 1 Time ($\mathrm{s}$) &${\color{gold}{5.90}}$  & $\mathbf{5.92}$    & $6.88$  & $6.46$ \\ 
 Lap 2 Time ($\mathrm{s}$) & ${\color{gold}{5.76}}$& $\mathbf{5.80}$   & $6.76$  & $6.18$ \\ 
 Lap 3 Time ($\mathrm{s}$) & ${\color{gold}{6.03}}$ & $\mathbf{6.04}$   &  $6.76$ & $6.14$ \\ 
 Violation Time ($\mathrm{s}$)&${\color{gold}{0.78}}$ & $\mathbf{0.64}$ & $2.42$  & $1.81$ \\
 Mean Deviation ($\mathrm{m}$)& ${\color{gold}{0.04}}$& $\mathbf{0.04}$  & $0.08$ & $0.07$ \\
 \hline
 \multicolumn{5}{c}{\bf Experiment 2} \\ 
 \hline
 Lap 1 Time ($\mathrm{s}$) &   ${\color{gold}{5.90}}$& $\mathbf{5.90}$   &  $5.92$   & $6.46$ \\ 
 Lap 2 Time ($\mathrm{s}$) &  ${\color{gold}{5.69}}$ & ${5.74}$    &  $\mathbf{5.68}$  & $6.38$ \\ 
 Lap 3 Time ($\mathrm{s}$) &  ${\color{gold}{6.70}}$ & $\mathbf{6.68}$    &  $6.88$  & $7.14$ \\ 
 Violation Time ($\mathrm{s}$)&  ${\color{gold}{0.45}}$& $\mathbf{0.44}$    &  $0.70$  & $3.10$ \\
 Mean Deviation ($\mathrm{m}$)& ${\color{gold}{0.04}}$ &  $\mathbf{0.04}$    &  $\mathbf{0.04}$ &  $0.08$  \\
 \hline
 \multicolumn{5}{c}{\bf Experiment 3} \\ 
 \hline
 Lap 1 Time ($\mathrm{s}$)&   ${\textcolor{gold}{5.94}}$& $\mathbf{6.00}$    &  $6.04$ & $6.54$ \\ 
 Lap 2 Time ($\mathrm{s}$)&  ${\textcolor{gold}{6.70}}$ &  $\mathbf{6.64}$   &  $7.02$  & $7.22$ \\ 
 Lap 3 Time ($\mathrm{s}$)&   ${\textcolor{gold}{6.66}}$& $\mathbf{6.78}$   &  $6.84$  & $6.79$ \\ 
 Violation Time ($\mathrm{s}$)& ${\textcolor{gold}{0.24}}$ & $0.27$    &  $\mathbf{0.25}$  & ${2.34}$\\
 Mean Deviation ($\mathrm{m}$)& ${\textcolor{gold}{0.04}}$ &  $0.06$   &   $\mathbf{0.04}$  & $0.08$ \\
 \hline
  {\bf Avg. Comput. Time} ($\mathrm{s}$)&${\textcolor{gold}{0.02}}$   &  $\mathbf{0.03}$        & $\mathbf{0.03}$ & $0.06$ \\
 \hline
  \end{tabular}
 \label{table: ETHZMobil Track}
  \vspace{-6mm}
\end{table}

\subsubsection{Gradual friction decay (Experiment 1)} 
This scenario represents a ubiquitous challenge in racing where friction conditions deteriorate progressively throughout the race. In competitive racing, gradual friction decay occurs due to tire wear from cumulative lap loading, rising tire temperatures causing compound breakdown, evolution of track surface conditions as rubber is deposited, and gradual weather changes. These progressive changes require continuous adaptation. In this experiment, we consider a linear decay of the actual friction coefficient $\mu$ by 2\% per second. This tests each controller's ability to track and adapt to slowly evolving dynamics—a scenario where traditional controllers maintain acceptable performance initially but gradually degrade.

\begin{figure*}
\vspace{-5mm}
    \centering
    \subfloat[LLA-MPC ($N=20000$)]{
         \includegraphics[width=.3\linewidth]{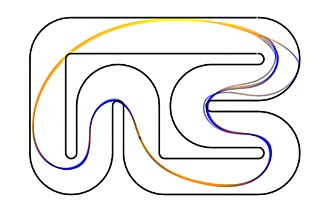}}
   \subfloat[APACRace]{
        \includegraphics[width=.3\linewidth]{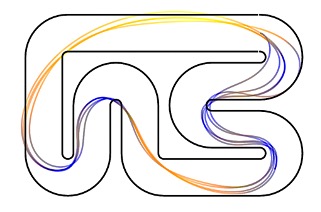}}
    \subfloat[Oracle]{
    \includegraphics[width=.29\linewidth]{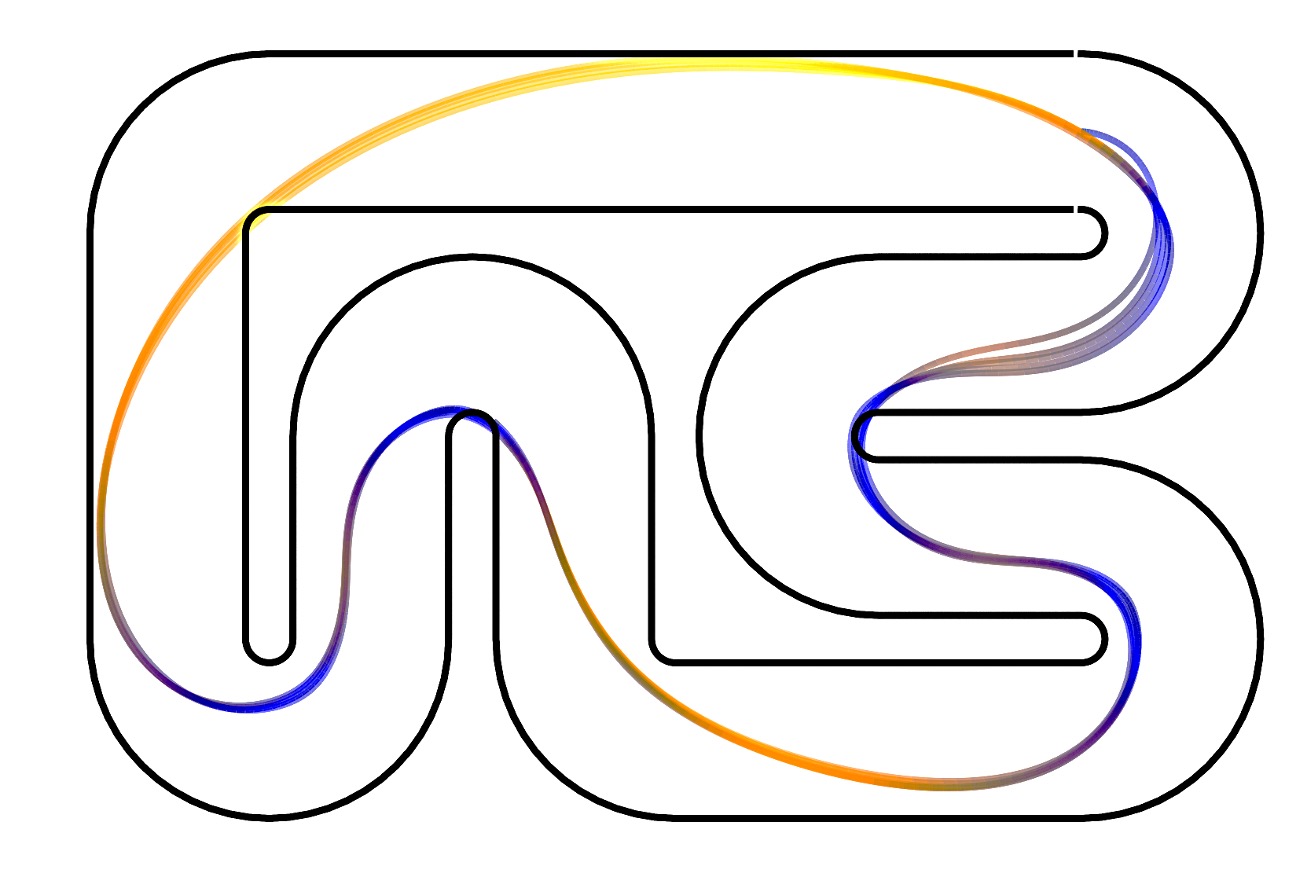}}
    \subfloat{
    \includegraphics[width=0.06\linewidth]{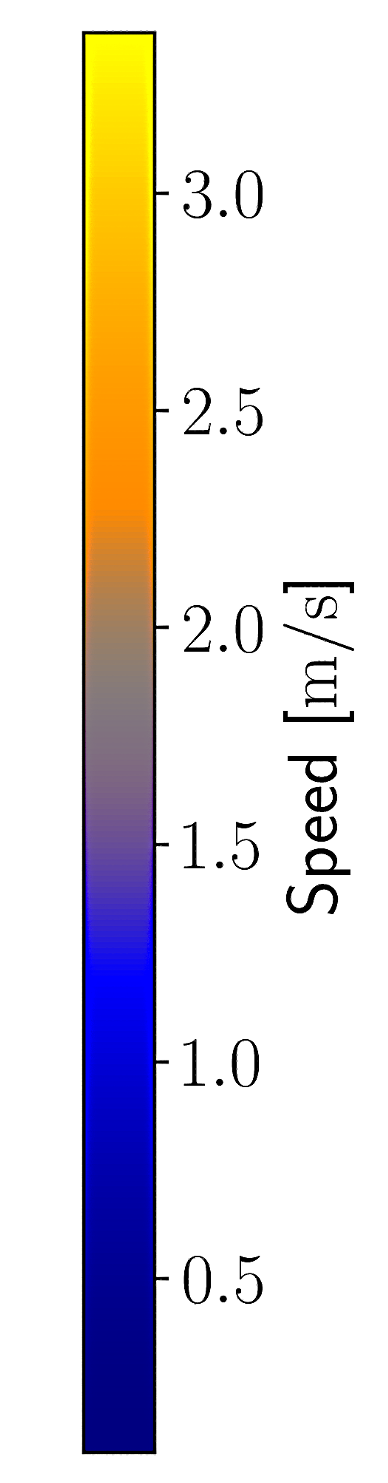}}
    \caption{{\bf(Experiment 1)} Comparison of racing trajectories on the ETHZMobil track under progressive friction decay.}
   \label{fig:T2-CASE 2 (GRAD AFTER)}
   \vspace{-5mm}
\end{figure*}

As can be seen in \autoref{table: ETHZ Track} and \autoref{table: ETHZMobil Track}, our method outperforms APACRace for all three metrics. \autoref{fig:all mus} (top) shows the different methods' ability to estimate the friction coefficient on the ETHZ track. LLA-MPC and APACRace perform similarly in this scenario, with LLA-MPC converging faster due to its learning-free nature. APACRace needs to complete its initial training phase before it can start the estimation. \autoref{fig:T2-CASE 2 (GRAD AFTER)} illustrates the different methods' resultant trajectories on the ETHZMobil track. As can be seen, LLA-MPC successfully adapts to the gradual loss of grip, mimicking the Oracle with high speeds.

% \begin{figure*}
%     \centering
%     \subfloat[LLA-MPC]{
%          \includegraphics[trim={0 0 260 0},clip,width=0.27\linewidth]{results/LLA T2/CASE 2 (GRAD AFTER)/Traj_Velocity_c2.png}}
%    \subfloat[APACRace]{
%         \includegraphics[trim={0 0 260 0},clip,width=0.27\linewidth]{results/APACRace T2/CASE 2 (GRAD AFTER)/Traj_Velocity_c2.png}}
%     \subfloat[Oracle]{
%     \includegraphics[trim={0 0 0 0},clip,width=0.363\linewidth]{results/GT T2/CASE 2 (GRAD AFTER)/Traj_Velocity2.png}}
%     \caption{{\bf(Experiment 1)} Comparison of racing trajectories on the ETHZMobil track under progressive friction decay with speed shown by a color map.}
%    \label{fig:T2-CASE 2 (GRAD AFTER)}
% \end{figure*}

\subsubsection{Sudden Friction Drop (Experiment 2)}
This scenario represents a critical challenge in racing, where surface properties change abruptly, demanding immediate adaptation. In real-world racing, sudden friction changes can occur due to transitions from dry to wet pavement during localized rain, patches of oil or debris on the track, forced deviations from the racing line during overtakes, or shifts between different surface materials such as concrete and asphalt. These unpredictable variations often leave little to no time for gradual adaptation, posing a severe test for control strategies. To simulate this challenge, we introduce a sudden $40\%$ reduction in the friction coefficient after completing the 1st lap. This scenario evaluates each controller’s ability to handle abrupt, unforeseen changes without prior adaptation time. Traditional learning-based approaches, which rely on gradual updates, struggle under such conditions. In contrast, our method’s ability to instantly select from a set of pre-computed models provides a significant advantage, enabling rapid adaptation. Note that by the beginning of the 2nd lap, the approach proposed in \cite{kalaria2024adaptive} completes its training phase and is supposed to give its peak adaptation performance. 
% shift between ... ?

\begin{figure*}
    \centering
    \subfloat[LLA-MPC ($N=20000$)]{
         \includegraphics[trim={0 0 260 0},clip,width=0.23\linewidth]{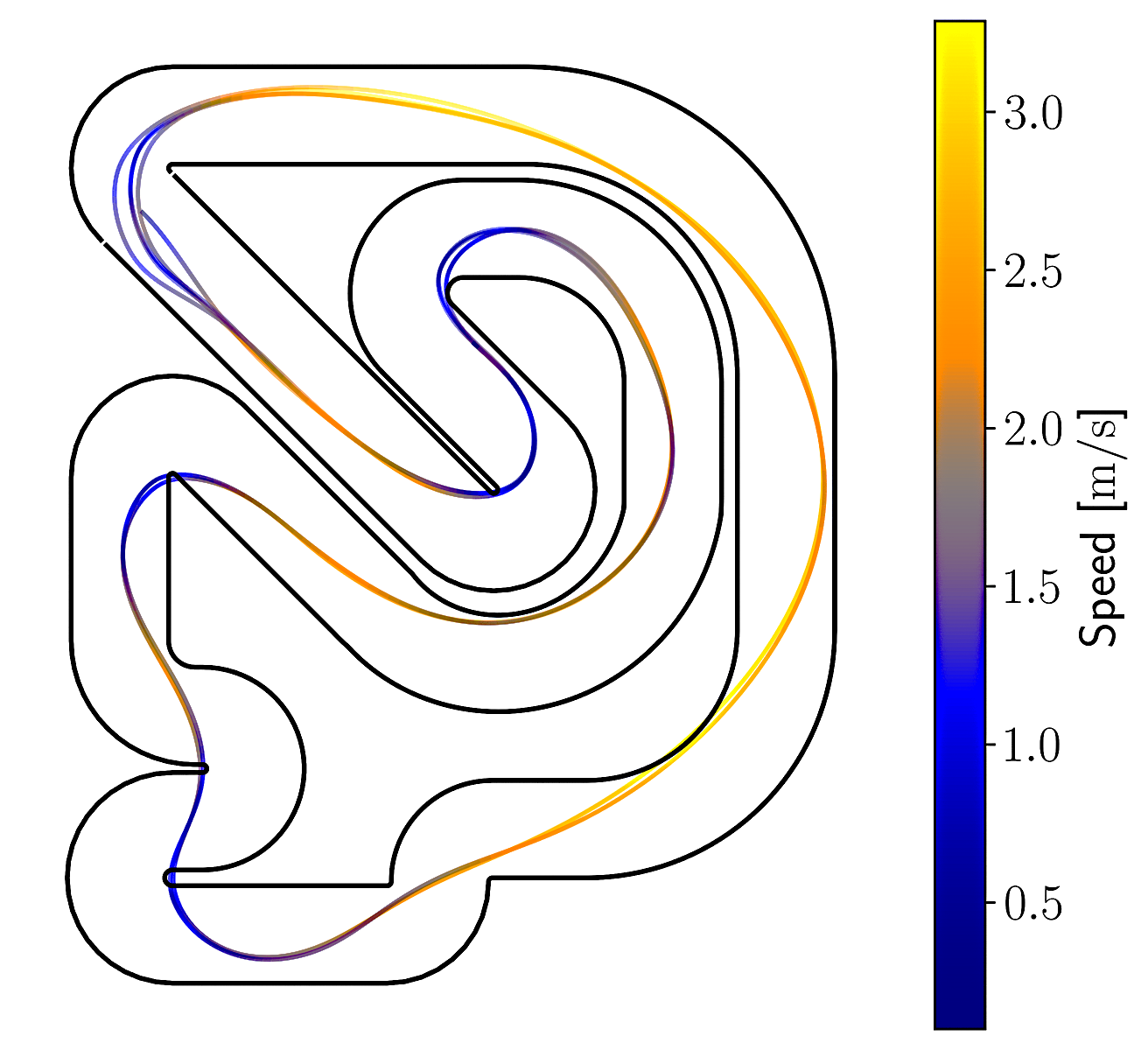}}
    \subfloat[APACRace]{
        \includegraphics[trim={0 0 260 0},clip,width=0.23\linewidth]{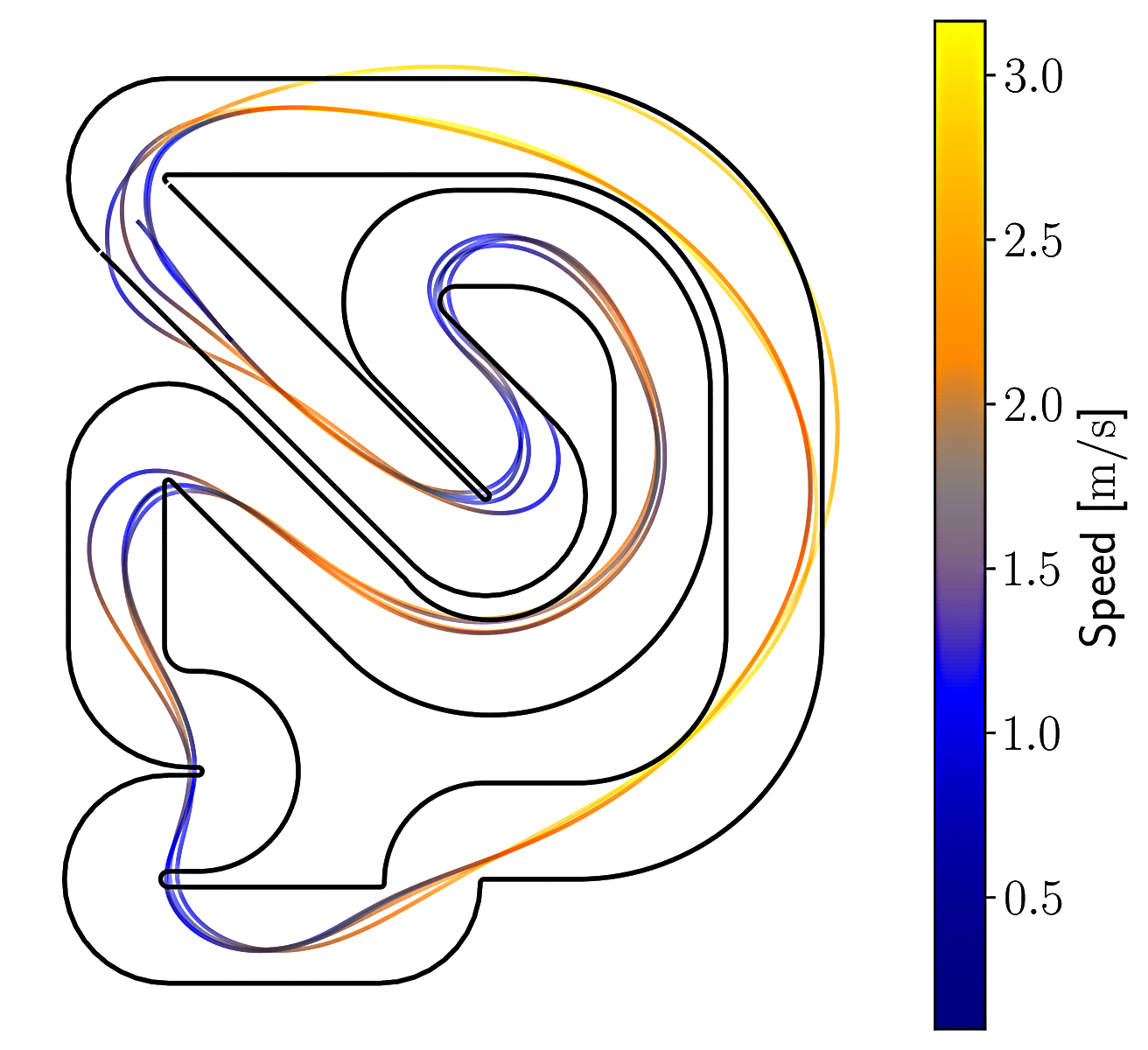}}
    \subfloat[Oracle]{
    \includegraphics[trim={0 0 0 0},clip,width=0.31\linewidth]{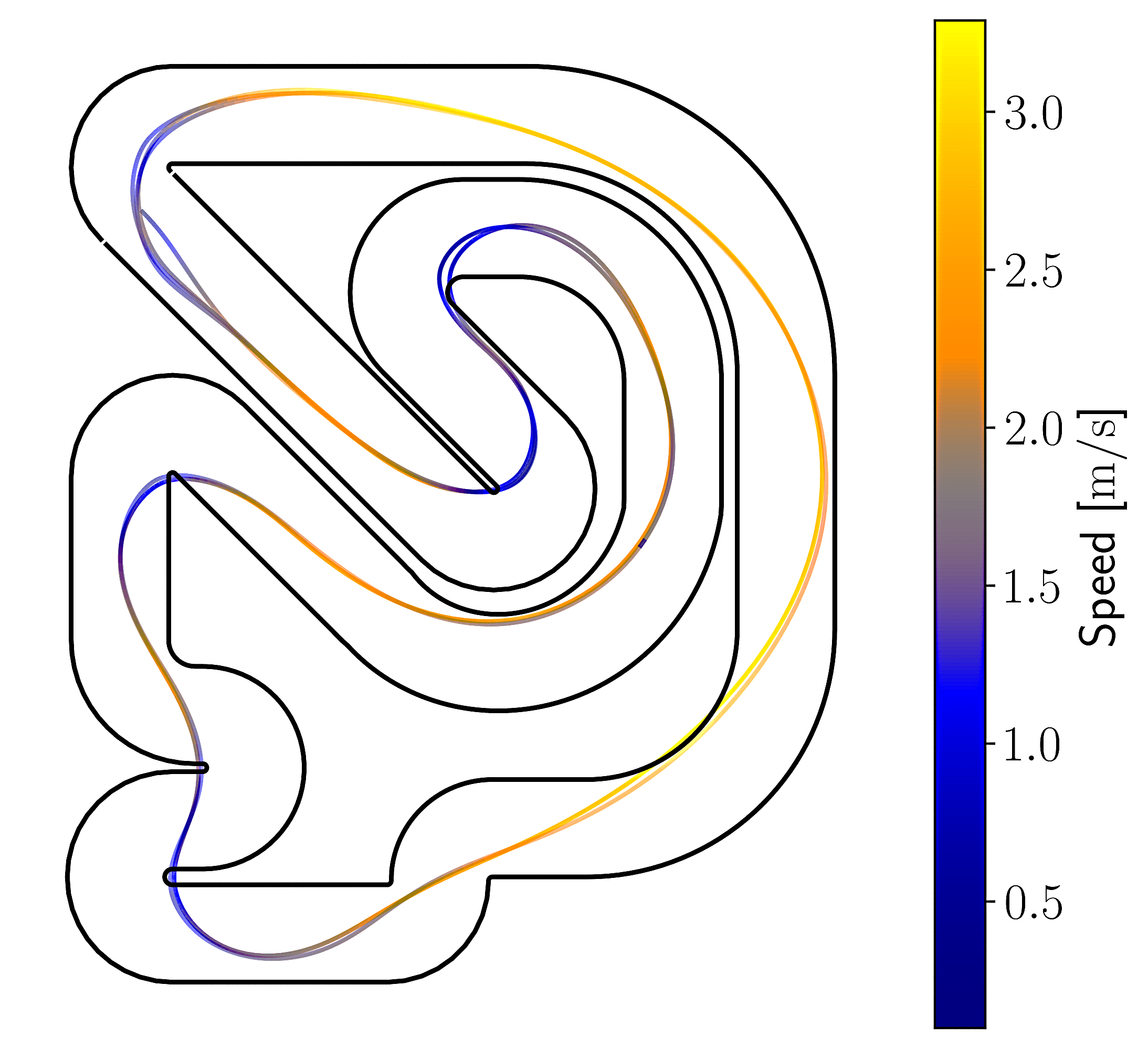}}
    \caption{{\bf(Experiment 2)} Comparison of racing trajectories under a sudden friction drop on the ETHZ track.}
    \label{fig: CASE 4 (SUDD AFTER)}
    \vspace{-5mm}
\end{figure*}

As shown in \autoref{table: ETHZ Track} and \autoref{table: ETHZMobil Track}, our method outperforms APACRace for all three metrics. \autoref{fig:all mus} (middle) shows the different methods' ability to estimate the friction coefficient in the ETHZ track. LLA-MPC converges very fast to the actual friction coefficient at the beginning and stays very close to it, and it adapts rapidly to the sudden change and stays very close to the actual friction coefficient afterward. On the other hand, APACRace starts with a larger error even after the training phase and responds very slowly to the sudden change. \autoref{fig: CASE 4 (SUDD AFTER)} demonstrates the impact of a sudden friction drop on the ETHZ track. LLA-MPC successfully maintains higher speeds while staying on track, closely following the Oracle. In contrast, APACRace struggles with the abrupt change, leading to off-track excursions.

\subsubsection{Early Sudden Friction Drop (Experiment 3)}
We repeat the previous experiment but introduce the sudden 
40\% friction reduction earlier, \textit{within the first lap}, before APACRace completes its training phase. This scenario further increases the challenge, as controllers must adapt immediately without the benefit of prior learning. While LLA-MPC leverages its diverse bank of models to handle the abrupt grip loss effectively, APACRace struggles due to its reliance on progressive learning, leading to performance degradation.

\begin{figure*}
    \centering
    \subfloat[LLA-MPC ($N=20000$)]{
         \includegraphics[trim={0 0 260 0},clip,width=0.23\linewidth]{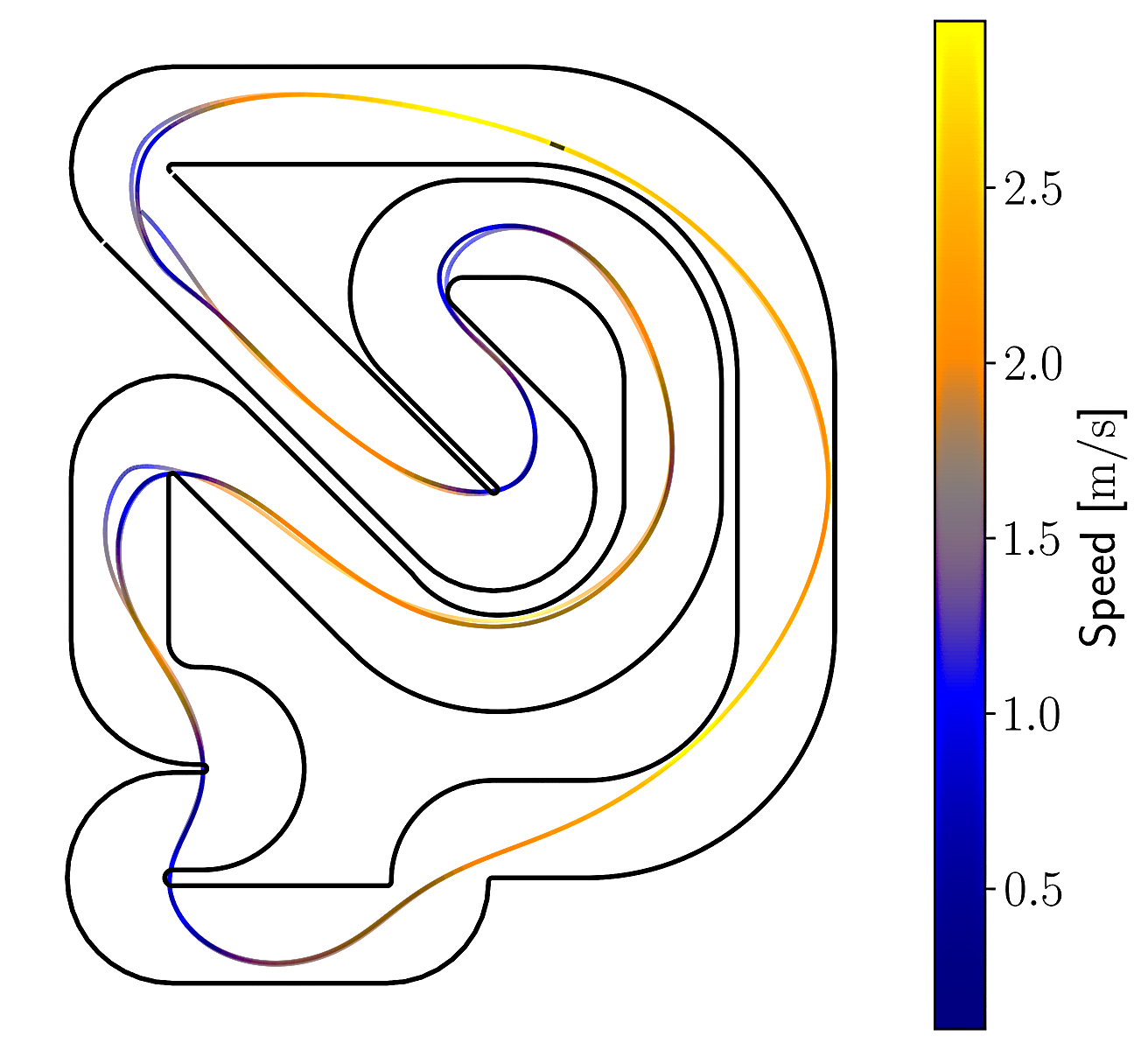}}
    \subfloat[APACRace]{
        \includegraphics[trim={0 0 260 0},clip,width=0.23\linewidth]{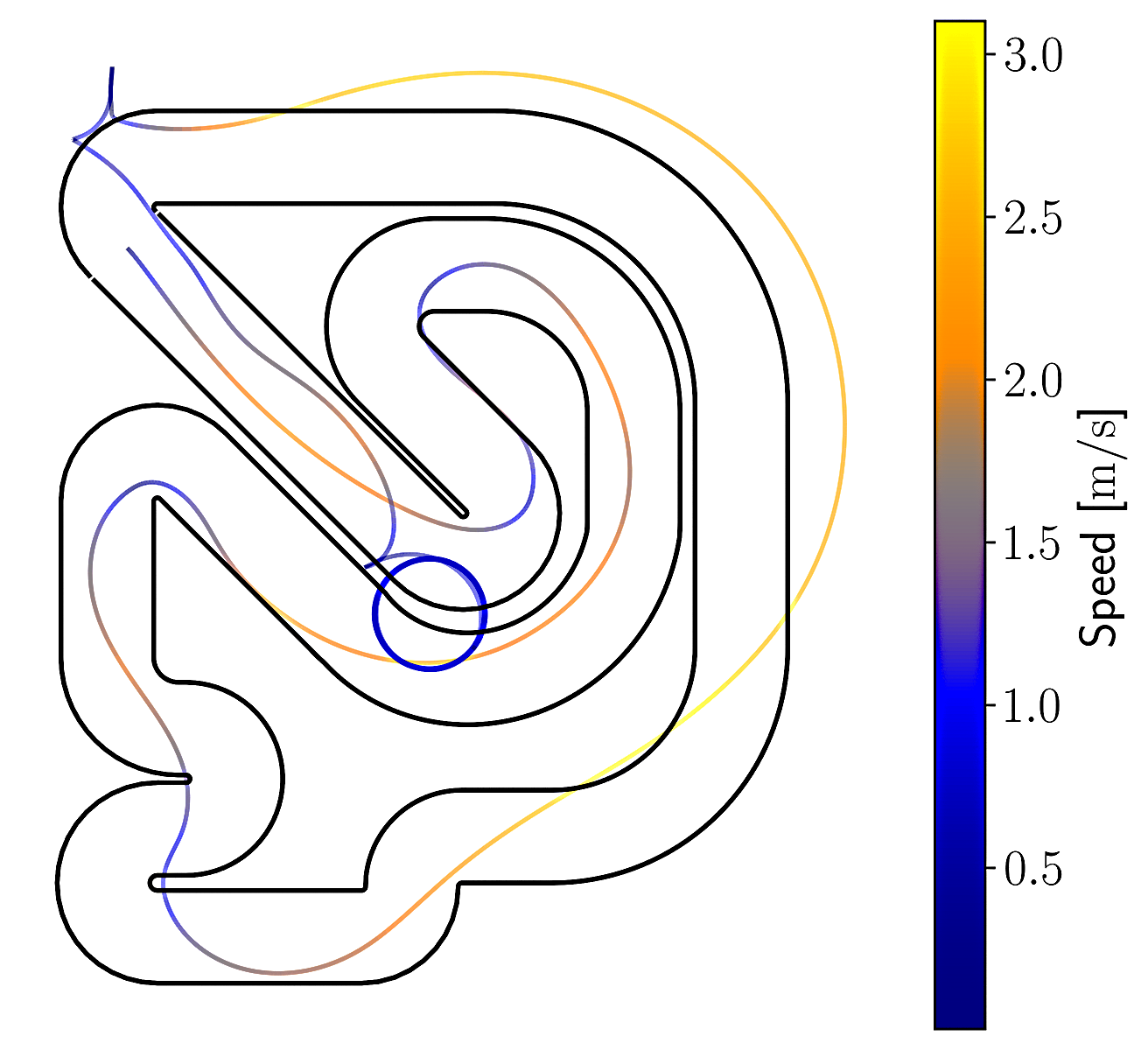}}
    \subfloat[Oracle]{
    \includegraphics[trim={0 0 0 0},clip,width=0.31\linewidth]{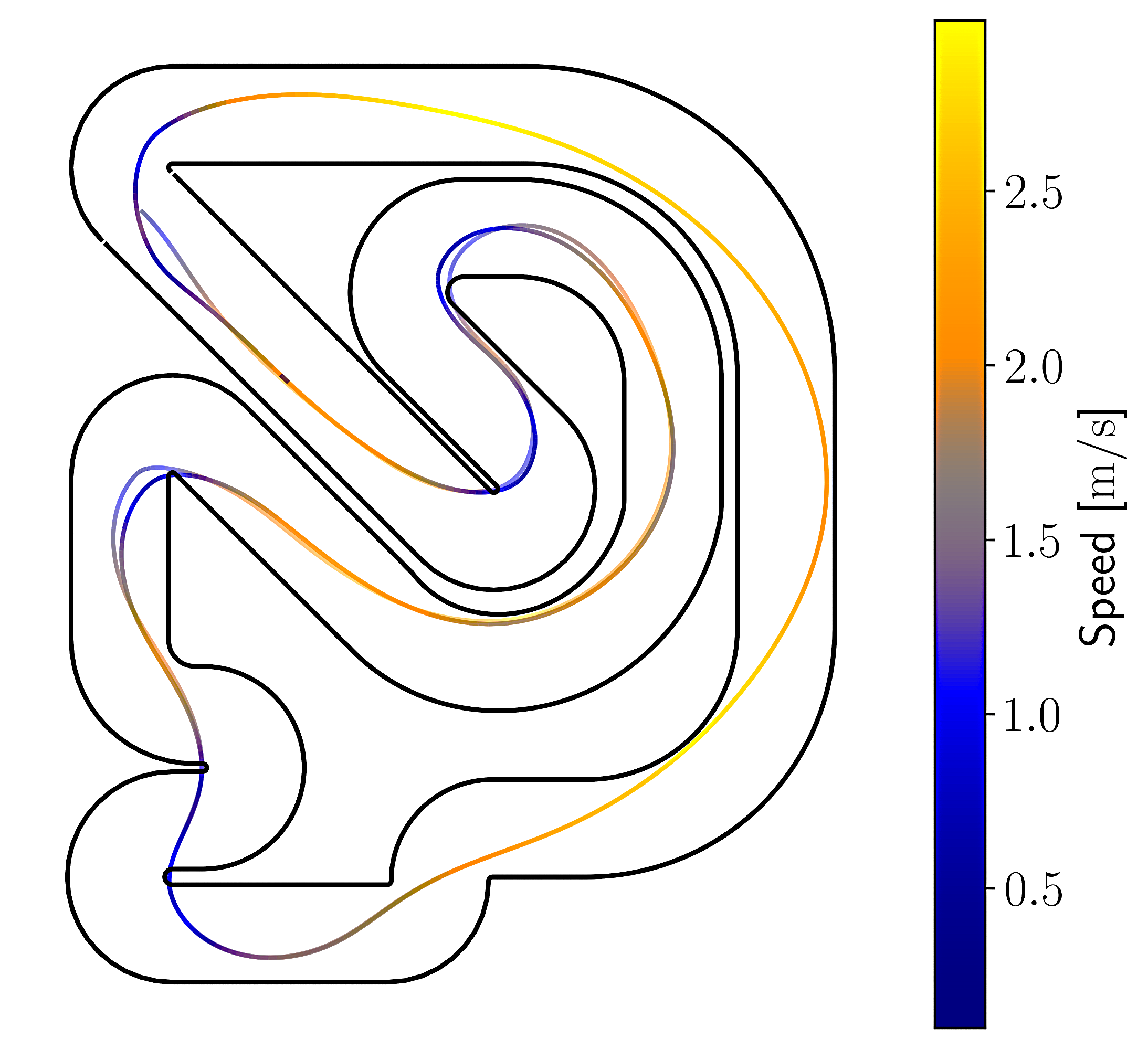}}
    \caption{{\bf(Experiment 3)} Comparison of racing trajectories under an \textbf{early} sudden friction drop on the ETHZ track.}
    \label{fig: CASE 3 (SUDD BEG)}
    \vspace{-5mm}
\end{figure*}

As shown in \autoref{table: ETHZ Track} and \autoref{table: ETHZMobil Track}, our proposed method outperforms APACRace for all three metrics except for the violation time in the 3rd lap on the ETHZMobil track. In the ETHZ track, we can see that APACRace was not able to complete the 2nd and 3rd laps and went out of track several times, making it not possible to compute its violation time and mean deviation.  \autoref{fig:all mus} (bottom) shows the different methods' ability to estimate the friction coefficient in the ETHZ track. Similar to experiment 2, LLA-MPC converges very fast to the actual friction coefficient, and it adapts rapidly to the sudden change. APACRace completely misses the true friction coefficient after completing its training phase. \autoref{fig: CASE 3 (SUDD BEG)} demonstrates the impact of a sudden friction drop on the ETHZ track. As can be seen, LLA-MPC's performance was not affected much by the earlier sudden change. In contrast, APACRace was completely interrupted by this early sudden change, making control infeasible.

The results of all experiments highlight LLA-MPC's superior adaptability and stability, particularly in handling sudden friction changes. In all experiments, LLA-MPC consistently achieves lower lap times and reduced mean deviation compared to APACRace, demonstrating its ability to maintain optimal racing trajectories. Additionally, APACRace exhibits significantly higher violation times, particularly in Experiment 3 (early sudden drop in friction), where it struggles to stay within track boundaries. The trends are consistent across both tracks, reinforcing the effectiveness of LLA-MPC in adapting to varying conditions while maintaining competitive lap times and minimizing deviations.

\subsection{CARLA Simulations}
For the high-fidelity simulator CARLA, we perform an experiment similar to experiment 1 in the previous section. Specifically, we let the actual friction coefficient decay linearly. Our results show that our method is responsive to changes in the environment and that our method can operate in real-time. Due to space limitation and since CARLA provides rendered videos, we present the results of the CARLA simulations on the paper's website: \href{https://github.com/DRIVE-LAB-CMU/LLA-MPC}{https://github.com/DRIVE-LAB-CMU/LLA-MPC}. %\textcolor{red}{TODO: Add a figure and some discussion}. 

% \vspace{-2mm}
\section{Conclusion and Future Work}
\label{sec:conclusion}
We introduced LLA-MPC, a novel framework for real-time adaptation in autonomous racing. By leveraging a model bank and integrating look-back and look-ahead mechanisms, LLA-MPC achieves immediate adaptation without requiring prior training or data collection. Our approach ensures fast and robust performance across diverse track conditions, handling both gradual and sudden friction variations with rapid responsiveness. Experimental results demonstrate that LLA-MPC significantly outperforms state-of-the-art adaptive control methods in both adaptation speed and trajectory tracking in high-speed racing scenarios. Its learning-free nature and computational efficiency make it well-suited for real-world deployment, eliminating the constraints of data-intensive approaches.  

Future work will explore enhancing LLA-MPC by integrating multiple model representations, including deep learning-based approaches, and improving sim-to-real transfer for deployment in real-world hardware. Additionally, we plan to investigate hybrid optimization techniques, such as cross-entropy methods, particle filtering or genetic algorithms, to refine model selection and improve adaptation efficiency by dynamically adjusting the sampling distribution.

% \vspace{-2mm}

\section{Acknowledgment}
We thank the author of APACRace, Dvij Kalaria, for the insightful discussions about his work and implementation.
% \vspace{-4mm}
\printbibliography

@article{betz2022autonomous,
  title={Autonomous vehicles on the edge: A survey on autonomous vehicle racing},
  author={Betz, Johannes and Zheng, Hongrui and Liniger, Alexander and Rosolia, Ugo and Karle, Phillip and Behl, Madhur and Krovi, Venkat and Mangharam, Rahul},
  journal={IEEE Open Journal of Intelligent Transportation Systems},
  volume={3},
  pages={458--488},
  year={2022},
  publisher={IEEE}
}

@article{narendra1997adaptive,
  title={Adaptive control using multiple models},
  author={Narendra, Kumpati S and Balakrishnan, Jeyendran},
  journal={IEEE transactions on automatic control},
  volume={42},
  number={2},
  pages={171--187},
  year={1997},
  publisher={IEEE}
}

@book{maybeck1982stochastic,
  title={Stochastic models, estimation, and control},
  author={Maybeck, Peter S},
  volume={3},
  year={1982},
  publisher={Academic press}
}

@article{liniger2015optimization,
  title={Optimization-based autonomous racing of 1: 43 scale RC cars},
  author={Liniger, Alexander and Domahidi, Alexander and Morari, Manfred},
  journal={Optimal Control Applications and Methods},
  volume={36},
  number={5},
  pages={628--647},
  year={2015},
  publisher={Wiley Online Library}
}

@article{bakker1987tyre,
  title={Tyre modelling for use in vehicle dynamics studies},
  author={Bakker, Egbert and Nyborg, Lars and Pacejka, Hans B},
  journal={SAE transactions},
  pages={190--204},
  year={1987},
  publisher={JSTOR}
}

@article{heilmeier2020minimum,
  title={Minimum curvature trajectory planning and control for an autonomous race car},
  author={Heilmeier, Alexander and Wischnewski, Alexander and Hermansdorfer, Leonhard and Betz, Johannes and Lienkamp, Markus and Lohmann, Boris},
  journal={Vehicle System Dynamics},
  year={2020},
  publisher={Taylor \& Francis}
}

@inproceedings{dosovitskiy2017carla,
  title={CARLA: An open urban driving simulator},
  author={Dosovitskiy, Alexey and Ros, German and Codevilla, Felipe and Lopez, Antonio and Koltun, Vladlen},
  booktitle={Conference on robot learning},
  pages={1--16},
  year={2017},
  organization={PMLR}
}

@inproceedings{jain2021bayesrace,
  title={BayesRace: Learning to race autonomously using prior experience},
  author={Jain, Achin and O’Kelly, Matthew and Chaudhari, Pratik and Morari, Manfred},
  booktitle={Conference on Robot Learning},
  pages={1918--1929},
  year={2021},
  organization={PMLR}
}

@inproceedings{hewing2018cautious,
  title={Cautious nmpc with gaussian process dynamics for autonomous miniature race cars},
  author={Hewing, Lukas and Liniger, Alexander and Zeilinger, Melanie N},
  booktitle={2018 European control conference (ECC)},
  pages={1341--1348},
  year={2018},
}

@article{rosolia2019learning,
  title={Learning how to autonomously race a car: a predictive control approach},
  author={Rosolia, Ugo and Borrelli, Francesco},
  journal={IEEE Transactions on Control Systems Technology},
  volume={28},
  number={6},
  pages={2713--2719},
  year={2019},
  publisher={IEEE}
}

@inproceedings{becker2023model,
  title={Model-and acceleration-based pursuit controller for high-performance autonomous racing},
  author={Becker, Jonathan and Imholz, Nadine and Schwarzenbach, Luca and Ghignone, Edoardo and Baumann, Nicolas and Magno, Michele},
  booktitle={2023 IEEE International Conference on Robotics and Automation (ICRA)},
  pages={5276--5283},
  year={2023},
  organization={IEEE}
}

@article{voser2010analysis,
  title={Analysis and control of high sideslip manoeuvres},
  author={Voser, Christoph and Hindiyeh, Rami Y and Gerdes, J Christian},
  journal={Vehicle System Dynamics},
  volume={48},
  number={S1},
  pages={317--336},
  year={2010},
  publisher={Taylor \& Francis}
}

@inproceedings{raji2022motion,
  title={Motion planning and control for multi vehicle autonomous racing at high speeds},
  author={Raji, Ayoub and Liniger, Alexander and Giove, Andrea and Toschi, Alessandro and Musiu, Nicola and Morra, Daniele and Verucchi, Micaela and Caporale, Danilo and Bertogna, Marko},
  booktitle={2022 IEEE 25th International Conference on Intelligent Transportation Systems (ITSC)},
  pages={2775--2782},
  year={2022},
  organization={IEEE}
}

@article{seong2023model,
  title={Model parameter identification via a hyperparameter optimization scheme for autonomous racing systems},
  author={Seong, Hyunki and Chung, Chanyoung and Shim, David Hyunchul},
  journal={IEEE Control Systems Letters},
  volume={7},
  pages={1652--1657},
  year={2023},
  publisher={IEEE}
}

@book{rajamani2011vehicle,
  title={Vehicle dynamics and control},
  author={Rajamani, Rajesh},
  year={2011},
  publisher={Springer Science \& Business Media}
}

@inproceedings{brunner2017repetitive,
  title={Repetitive learning model predictive control: An autonomous racing example},
  author={Brunner, Maximilian and Rosolia, Ugo and Gonzales, Jon and Borrelli, Francesco},
  booktitle={2017 IEEE 56th annual conference on decision and control (CDC)},
  pages={2545--2550},
  year={2017},
  organization={IEEE}
}

@article{bodmer2024optimization,
  title={Optimization-based system identification and moving horizon estimation using low-cost sensors for a miniature car-like robot},
  author={Bodmer, Sabrina and Vogel, Lukas and Muntwiler, Simon and Hansson, Alexander and Bodewig, Tobias and Wahlen, Jonas and Zeilinger, Melanie N and Carron, Andrea},
  journal={arXiv preprint arXiv:2404.08362},
  year={2024}
}

@article{kabzan2019learning,
  title={Learning-based model predictive control for autonomous racing},
  author={Kabzan, Juraj and Hewing, Lukas and Liniger, Alexander and Zeilinger, Melanie N},
  journal={IEEE Robotics and Automation Letters},
  volume={4},
  number={4},
  pages={3363--3370},
  year={2019},
  publisher={IEEE}
}

@article{dikici2025learning,
  title={Learning-Based On-Track System Identification for Scaled Autonomous Racing in Under a Minute},
  author={Dikici, Onur and Ghignone, Edoardo and Hu, Cheng and Baumann, Nicolas and Xie, Lei and Carron, Andrea and Magno, Michele and Corno, Matteo},
  journal={IEEE Robotics and Automation Letters},
  year={2025},
  publisher={IEEE}
}

@inproceedings{kalaria2024adaptive,
  title={Adaptive planning and control with time-varying tire models for autonomous racing using extreme learning machine},
  author={Kalaria, Dvij and Lin, Qin and Dolan, John M},
  booktitle={2024 IEEE International Conference on Robotics and Automation (ICRA)},
  pages={10443--10449},
  year={2024},
  organization={IEEE}
}

@inproceedings{tsuchiya2024online,
  title={Online Adaptation of Learned Vehicle Dynamics Model with Meta-Learning Approach},
  author={Tsuchiya, Yuki and Balch, Thomas and Drews, Paul and Rosman, Guy},
  booktitle={2024 IEEE/RSJ International Conference on Intelligent Robots and Systems (IROS)},
  pages={802--809},
  year={2024},
  organization={IEEE}
}

@inproceedings{kalaria2025agile,
  title={Agile Mobility with Rapid Online Adaptation via Meta-learning and Uncertainty-aware MPPI},
  author={Kalaria, Dvij and Xue, Haoru and Xiao, Wenli and Tao, Tony and Shi, Guanya and Dolan, John M},
  booktitle={2025 International Conference on Robotics and Automation (ICRA)},
  year={2025}
}

@inproceedings{xiao2025anycar,
  title={Anycar to anywhere: Learning universal dynamics model for agile and adaptive mobility},
  author={Xiao, Wenli and Xue, Haoru and Tao, Tony and Kalaria, Dvij and Dolan, John M and Shi, Guanya},
  booktitle={2025 International Conference on Robotics and Automation (ICRA)},
  year={2025}
}

@article{nagy2023ensemblegaussianprocessesadaptive,
  title={Ensemble gaussian processes for adaptive autonomous driving on multi-friction surfaces},
  author={Nagy, Tom{\'a}{\v{s}} and Amine, Ahmad and Nghiem, Truong X and Rosolia, Ugo and Zang, Zirui and Mangharam, Rahul},
  journal={IFAC-PapersOnLine},
  volume={56},
  number={2},
  pages={494--500},
  year={2023},
  publisher={Elsevier}
}

% Hence, the key novelty is merging \emph{model-based stochastic optimal control} with a \emph{data-driven model selection} mechanism in order to tackle uncertain dynamics without an overly conservative single-model design. This yields an adaptive framework capable of robustly tracking complex trajectories despite parameter variations and modeling inaccuracies.

\end{document}